\documentclass{article}
 \usepackage{graphicx}
\usepackage{amsthm}
\usepackage{amsmath}
\usepackage{amssymb}

\title{The Use of Deep Learning for Symbolic
Integration \\
A Review of (Lample and Charton, 2019)}

\author{
Ernest Davis \\
Dept. of Computer Science \\ New York University \\ New York, NY 10012 \\
{\small davise@cs.nyu.edu}}

\begin{document}
\maketitle

\begin{abstract}
Lample and Charton (2019) describe a system that uses deep learning 
technology to compute symbolic, indefinite integrals, and to find symbolic
solutions to first- and second-order 
ordinary differential equations, when the solutions are elementary
functions. 
They found that, over a particular test set,
the system could find solutions more successfully than sophisticated packages
for symbolic mathematics such as Mathematica run with a long time-out.
This is an impressive accomplishment, as far as it goes. However,
the system can handle only a quite limited subset of the problems that
Mathematica deals with, and the test set has significant built-in biases. 
Therefore the claim that this outperforms 
Mathematica on symbolic integration needs to be very much qualified.

\end{abstract}

Lample and Charton (2019) describe a system (henceforth LC)
that uses deep learning 
technology to compute symbolic, indefinite integrals, and to find symbolic
solutions to first- and second-order 
ordinary differential equations, when the solutions are elementary
functions (i.e. compositions of the arithmetic operators with the exponential
and trigonometric functions and their inverses). They found that, over a 
particular test set,
LC could find solutions more successfully than sophisticated packages
for symbolic mathematics such as Mathematica given a long time-out.

This is an impressive accomplishment; however, it is important to understand
its scope and limits.

We will begin by discussing the case of symbolic integration, which is simpler.
Our discussion of ODE's is much the same; however, that introduces 
technical complications that are largely extraneous to the points we want to
make.

\section{Symbolic integration}
There are three categories of computational symbolic mathematics that are
important here:
\begin{itemize}
\item {\bf Symbolic differentiation.} Using the standard rules for 
differential calculus, this is easy to program and efficient to execute.

\item {\bf Symbolic integration}
This is difficult. In most cases, the integral of an elementary function
that is not extremely simple is not, itself, an elementary function. In
principle, the decision problem whether the integral of an elementary function 
is itself elementary is undecidable (Richardson, 1969).
Even
in the cases where the integral is elementary, finding it can be very
difficult. Nonetheless powerful modules for symbolic integration have been
incorporated in systems for symbolic math like Mathematica, Maple, and 
Matlab.

\item {\bf Simplification of symbolic expressions.}
The decision problem of determining whether an elementary expression is 
identically equal to zero is undecidable (Richardson, 1969).
Symbolic math platforms incorporate powerful modules, but
but building a high-quality system is a substantial undertaking.
\end{itemize}

If one can, in one way or another, conjecture that the integral of elementary 
function $f$ is elementary function $g$ (both functions being specified 
symbolically) then verifying that conjecture involves,
first computing the derivative $h = f^{\prime}$ and, second, determining that
the expression $h-g$ simplifies to 0. As we have stated, the first step is 
easy; the second step is hard in principle but often reasonably
straightforward in practice.

Given an elementary expression $f$, finding an elementary symbolic integral 
is, in general, a search in an enormous and strange state space for something
that most of the time does not even exist. Even if you happen to know that it 
exists, as is the case with the test examples used by Lample and Charton,
it remains a very hard problem.

\section{What LC does and how it works}
At a high level, LC works as follows:
\begin{itemize}
\item A large corpus of examples (80 million)
was created synthetically by generating random,
complex pairs of symbolic expressions and their derivatives. We will discuss
below how that was done.
\item A seq2seq transformer model is trained on the corpus.
\item At testing time, given a function $g$ to integrate, the model was 
executed, using a beam search of width either 1, 10, or 50.
An answer $f$ produced by the model
was checked using the procedure described above:
the symbolic differentiator was applied to $f$, and then the symbolic
simplifier tested whether $f'=g$. 
\end{itemize}

In effect, the process of integration is being treated as something like 
machine translation: the source is the integrand, the target is the 
integral. 

Three techniques were used to generate integral/derivative pairs:
\begin{itemize}
\item {\bf Forward generation} (FWD). Randomly generate a symbolic function; 
give it to a preexisting symbolic integrator; if it finds an answer, then 
record the pair. 20 million such pairs were created. This tends to generate 
pairs with comparatively small derivative and large integrals, in terms of
the size of the symbolic expression.
\item {\bf Backward generation} (BWD). Randomly generate a symbolic function;
compute its derivative; and record the pair. 40 million such pairs were 
created. The form of the derivative is simplified by symbolic techniques
before the pair is recorded.
This approach
tends to generate pairs with comparatively small integrals and
derivatives that are almost always much larger.
\item {\bf Integration by parts} (IBP). If, for some functions $f$ and $G$,
LC has computed that the
integral of $f$ is $F$ and that the integral of the product $fG=H$, then,
by the rule of integration by parts.
\[ \int Fg = FG - H \] 
where $g$ is the derivative of $G$. 
It can now record $Fg$ and its integral as a 
new pair. 20 million such pairs were created.
\end{itemize}

The comparisons to Mathematica, Maple, and Matlab were carried out using
entirely items generated by BWD,
They found that LC was able to solve
a much higher percentage of this test set than Mathematica, Maple, or Matlab
giving an extended time out to all of these.
Mathematica, the highest scoring of these, was able to solve 84\% of the 
problems in the test set, whereas, running with a beam size of 1, 
LC produced the correct solution 
98.4\% of the time.

\section{No integration without simplification!}
There are many problems where it is critical to simplify an integrand
before carrying out the process of integration.

Consider the following integral:

\begin{equation}
\int \sin^{2}(e^{e^x}) + \cos^{2}(e^{e^x}) \: dx 
\end{equation}

At first glance, that looks scary, but in fact it is just a ``trick question'' 
that some malevolent calculus teacher might pose. The integrand is identically
equal to 1, so the integral is $x+c$. 

Obviously, one can spin out examples of this kind to arbitrary complexity.
The reader might enjoy evaluating this (2)

\[
\int \sin(e^{x}+\frac{e^{2x}-1}{2\cos^{2}(\sin(x))-1)}) - 
\cos(\frac{(e^{x}+1)(e^{x}-1)}{\cos(2\sin(x))})\sin(e^{x}) -
\cos(e^{(x^{3}+3x^{2}+3x+1)^{1/3}-1})
\sin(\frac{e^{2x}-1}{1-2\sin^{2}(\sin(x))}) 
\]

or not. Anyway, rule-based symbolic systems can and do quite easily carry out a
sequence of transformations to do the simplification here

This raises two issues:
\begin{itemize}
\item It is safe to assume that few examples of this form, of any significant
complexity, were included in Lample and Charton's corpus. BWD and IBP cannot
possibly generate these. FWD could, in principle, but the integrand would have
to be generated at random, which is extremely improbable. Therefore, LC 
was not tested on them.

\item Could LC
have found a solution to such a problem if it were tested on one?
The answer to that question is certainly yes, if the test procedure
begins by applying a high quality simplifier and reduces the integrand to
a simple form. 

Alternatively, the answer is also yes if, with integral (1), 
LC proposes ``$f(x)=x$'' as a candidate solution 
then the simplifier verifies that, indeed, $f'$ is equal to the complex
integrand. It does seem rather unlikely that LC, operating on
the integrand in equation (1), would propose $f(x)=x$ as a candidate. If
one were to construct a problem 
where a mess like integral (2) has some moderately
complicated solution --- say, $\log(x^{2}/\sin(x))$ --- which, of course,
is easily done, the likelihood that LC will find it seems still
smaller; though certainly there is no way to know until you try.

\end{itemize}

Fran\c{c}ois Charton informs me (personal communication) that in fact 
LC did not do simplifications at this stage and therefore would not
have been able to solve these problem.

This is not a serious strike against their general methodology.
The problem is easily fixed; they can add easily add calls
to the simplifier at the appropriate steps, and they could automatically
generate examples of this kind for their corpus
by using an ``uglifier'' that turns simple
expressions into equivalent complicated ones. 
But the point is that there a class of problems whose solution inherently
requires a high quality simplifier, and which currently is not being tested.

One might object that problems of this form are very artificial. But the entire 
problem that LC addresses is very artificial. If there is any 
natural application that tends to generate lots of problems of integrating
novel complicated 
symbolic expressions with the property that a significant fraction of those
have integrals that are elementary functions, I should like to
hear about it. As far as I know, the problem is purely of mathematical 
interest.

Another situation where simplification is critical: 
Suppose that you take an integrand
produced by BWD and make some small change. Almost certainly, the new function
$f$ has no elementary integral. Now give it to LC. LC will produce an 
answer $g$,
because it always produces an answer, and that answer will be wrong, 
because there is no right answer. In a situation where you actually cared about
the integral of $f$, it would be undesirable to accept $g$ as an answer.
So you can add a check; you differentiate $g$ and check whether $g'=f$. But
now you are checking for the equivalence of two very complicated expressions,
and again you would need a very high-powered simplifier.

\section{Differential equations}
Lample and Charton have developed a very ingenious technique in which you
can input any elementary function $f(x,c)$ with a single occurrence 
of parameter $c$, and find a
an ODE whose solution is $f(x,c)$ where $c$ is the free parameter of the
integral. They also can do the corresponding thing for second-order 
equations.

Their overall procedure was then essentially the same as for integrals:
They generated a large corpus of pairs of equations and solutions, 
and trained a seq2seq neural network. At testing time, LC used the neural
network to carry out a beam search which generated candidates; each 
candidate was tested to see whether it was a solution to the problem. 

Here the results were more mixed. With first-order equations, LC with a 
beam size of 1 comes out slightly ahead of Mathematica (81.2\% to 77.2\%)
with a beam size of 50, it comes out well ahead (97\%). With second
order equations, LC with a beam size of 1 does not do as well as Mathematica 
(40.8\% to 61.6\%) but with a beam size of 50 it attains 81.0\%.

The concerns that we have raised in the context of integration apply
here as well, suitably adapted.

\section{Special functions}
Systems such as Mathematica, Maple, and Matlab are able to solve symbolically
many symbolic integration problems and many differential equations in which
the solution is a special function (i.e. a non-elementary function with 
a standard name). For instance Mathematica can 
integrate the function $\log(1-x/x)$ to get the answer $-$PolyLog(2,x),
It can solve the equation
\[ x^{2}+y^{\prime \prime}(s) + xy^{\prime}(x) +(x^{2}-16)y(x) = 0 \]
to find the solution
$y(x) = c_{1}\mbox{BesselJ}(4,x) + c_{2}\mbox{BesselY}(4,x)$

In principle, LC could be extended to handle these.
In FWD, it would be a matter of including the pairs where the automated
integrator being called generates an expression with a special function.
In BWD, it would be a matter of generating expressions with special
functions and computing their derivatives.

With integration, the impact on performance might be small;
special functions that are integrals of elementary functions
are mostly unary 
(though PolyLog is binary), and therefore have only a moderate 
impact on the size of the state space. But the ODE solver is a different matter;
many of the functions that arise in solving
ODEs, such as the many variants of Bessel functions, are binary, and
adding expressions that include these expands the search space exponentially.
To put the point another way: When the ODE solver was tested,
Mathematica 
was searching through a space of solutions that are includes the
special functions, whereas LC was limited to the much smaller
space of the elementary functions. The tests were designed so that the
solution was always in the smaller space.
LC thus had an entirely unfair advantage.

\section{The Test Set}

There are also issues with the test set. The comparison with
Mathematica, Matlab, and Maple used a test set consisting entirely of problems
generated by BWD (problems generated by FWD by definition can be solved by 
symbolic integrators). These inevitably tend to have comparatively small
integrals (in expression size) and long integrals. Unless you are very lucky,
or unless an expression is full of addition and subtraction, the derivative 
of an expression of size $n$ has length 
$\Omega(n^2)$. For example the derivative 
of the function $\sin(\sin(\sin(\sin(x))))$ is 

\[
\cos(\sin(\sin(\sin(x)))) \cdot \cos(\sin(\sin(x))) \cdot \cos(\sin(x)) 
\cdot \cos(x)
\]
And in fact the average length of an integrand in the test set was 70 symbols
with a standard deviation of 47 symbols; thus, a large fraction of the test 
examples had 120 symbols or so. (Table 1 of Lample and Charton).

So what the comparison with Mathematica establishes is that, 
given a really long expression, which happens
to have a much shorter, exact symbolic integral, LC is awfully good at 
finding it. But that is a really special class. One can certainly understand
why the teams building Mathematica and so on have not considered 
this niche category of problem much of a priority. 

Another point that does not seem to have been tested is whether LC may have
been picking up on arbitrary artifacts of the differentiation process, such
as the order in which parts of a derivative are presented. For instance, the
derivative of a three level composed function $f(g(h(x)))$ is a product of
three terms $h'(x) \cdot g'(h(x)) \cdot f'(g(h(x))$. Any particular 
symbolic differentiator will probably generate these in a fixed order, such
as the one above. This particular choice of orderings will then be 
consistent thoughout the corpus, so LC will be trained and then tested
only with this ordering. A system like LC may have much more difficulty
finding the integral if the multiplicands are presented in any of the five
other possible orders.

The techniques that LC learns from BWD and FWD are very different. If LC is
trained only on BWD and tested on problems in FWD, 
then running with a beam size of 1, it finds the correct solution
to a problem in FWD only 18.9\% of the time; with a beam size of 50, the 
correct solution is among its top 50 candidates only 27.5\% of the time. 
Training it only on problems in FWD and testing it on problems in BWD it does
even worse, with corresponding success rates of 10.9\% and 17.2\%.

The procedure for generating corpus example will not succeed in creating
examples that combine features from FWD and BWD. For instance, if $f', f$
is a pair that would be naturally generated by FWD, and $g', g$ is a pair 
that would be naturally generated by BWD, then the sum $f'+g', f+g$ will
not be included in the corpus, and therefore will not be tested.

In fact: If one were to put together a test set of random, enormously complex 
integrands, LC would certainly give a wrong answer on nearly all of them, 
because only a small fraction would have an elementary integral. Mathematica,
certainly, would also fail to find an integral, but presumably it would not
give a wrong answer; it would either give up or time out. If you consider
that a wrong answer is worse than no answer, then on this test set, Mathematica
would beat LC by a enormous margin.

\section{Summary}
The fact that LC ``beat'' Mathematica on the test set of integration problems
produced by BWD is certainly impressive. But Lample and Charton's claim
\begin{quote}
[This] transformer model \ldots can perform extremely well both at computing
function integrals and and solving differential equations, outperforming \ldots
Matlab or Mathematica \ldots .
\end{quote}
is very much overstated, and requires significant qualification. The correct
statement, as regards integration, is as follows:

\begin{quote}
The transformer model outperforms Mathematica and Matlab in computing
symbolic 
indefinite integrals of enormously complex functions of a single variable `$x$'
whose integral is a much smaller 
elementary function containing no constant symbols
other than the integers $-5$ to 5.
\end{quote}

Since both BWD and FWD were limited to functions of a single
variable $x$, it is 
unknown whether LC can handle $\int t \: dt$ or $\int a \: dx$ (it's not clear
whether LC's input includes any way to specify the variable of integration) and
essentially certain that it cannnot handle $\int 1/(x^{2} + a^{2}) \: dx$. On 
problems like these, far from outperforming Mathematica and Matlab, it
falls far short of a high-school calculus student.

It is important to emphasize that
{\em the construction of LC is entirely dependent on
the pre-existing
symbolic processors developed over the last 50 years by experts in symbolic
mathematics.}  Moreover, as things now stand, extending LC to fill in some of
its gaps (e.g. the simplification problems described in section 3) would
make it even less of a stand-alone system and more dependent on 
conventional symbolic processors. There is no reason whatever to suppose
that NN-based systems will supercede symbolic mathematics systems any time
in the foreseeable future.

It goes without saying that LC has no understanding of the significance of
an integral or a derivative or even a function or a number. 
In fact, occasionally, it
outputs a solution that is not even a well-formed expression.  LC is like 
the worst possible student in a calculus class: it doesn't understand the 
concepts, it doesn't learned the rules, it has no idea what is the
significance of what it is doing, but it has looked at 80 million examples
and gotten a feeling of what integrands and their integrals look like. 

Finally LC, like the recent successes in
game-playing AI, depends on the ability to generate enormous quantities
of high-quality (in the case of LC, flawless) 
synthetic labelled data. In open-world domains, this is effectively
impossible. Therefore, the success of LC is in no way evidence that
deep learning or other such methods will suffice for high-level reasoning
in real-world situations.

\subsection*{Acknowledgements}
Thanks to Fran\c{c}ois Charton, Guillaume Lample, and
L\'{e}on Bottou for helpful discussions.

\subsection*{References}
Lample, Guillaume and Fran\c{c}ois Charton, 2019.
``Deep Learning for Symbolic Mathematics''
{\em NeurIPS-2019.}
https://arxiv.org/abs/1912.01412

Richardson, Daniel, 1969. ``Some undecidable problems involving elementary 
functions of a real variable." 
{\em The Journal of Symbolic Logic,} {\bf 33}:4, 514-520.

WolframAlpha, ``Integrals that Can and Cannot be Done'' \\
https://reference.wolfram.com/language/tutorial/IntegralsThatCanAndCannotBeDone.html

WolframAlpha, ``Symbolic Differential Equation Solving'' \\
https://reference.wolfram.com/language/tutorial/DSolveOverview.html

\end{document}